\documentclass[5p,twocolumn]{elsarticle}
\usepackage[utf8]{inputenc}
\usepackage{amsmath, amssymb}
\usepackage{multirow}
\usepackage{booktabs}
\usepackage{array}
\newcolumntype{P}[1]{>{\raggedright\arraybackslash}p{#1}}

\usepackage{makecell}
\usepackage{graphicx}
\setkeys{Gin}{keepaspectratio}
\usepackage{float}
\usepackage{minted}
\usepackage{placeins}
\usepackage{hyperref}
\hypersetup{
    colorlinks=true,
    linkcolor=blue,
    filecolor=magenta,      
    urlcolor=cyan,
    pdftitle={Overleaf Example},
    pdfpagemode=FullScreen,
}
\usepackage{xurl}
\usepackage{setspace}
\usepackage{booktabs}

\usepackage{braket}

\usepackage{appendix}

\usepackage{cleveref}
\usepackage{amsthm}
\newtheorem{theorem}{Theorem}

\title{Distribution-Guided and Constrained Quantum Machine Unlearning}
\author{Nausherwan Malik$^a$, Zubair Khalid$^a$, and Muhammad Faryad$^{b,*}$}
\address[]{Department of Electrical Engineering, Lahore University of Management Sciences, Lahore 54792, Pakistan.}
\address[]{Department of Physics, Lahore University of Management Sciences, Lahore 54792, Pakistan.\\
$^*$muhammad.faryad@lums.edu.pk}
\date{}

\begin{document}
\begin{abstract}
    Machine unlearning aims to remove the influence of specific training data from a learned model without full retraining. While recent work has begun to explore unlearning in quantum machine learning, existing approaches largely rely on fixed, uniform target distributions and do not explicitly control the trade-off between forgetting and retained model behaviour.
In this work, we propose a distribution-guided framework for class-level quantum machine unlearning that treats unlearning as a constrained optimization problem. Our method introduces a tunable target distribution derived from model similarity statistics, decoupling the suppression of forgotten-class confidence from assumptions about redistribution among retained classes. We further incorporate an anchor-based preservation constraint that explicitly maintains predictive behaviour on selected retained data, yielding a controlled optimization trajectory that limits deviation from the original model.
We evaluate the approach on variational quantum classifiers trained on the Iris and Covertype datasets. Results demonstrate sharp suppression of forgotten-class confidence, minimal degradation of retained-class performance, and closer alignment with the gold retrained model baselines compared to uniform-target unlearning. These findings highlight the importance of target design and constraint-based formulations for reliable and interpretable quantum machine unlearning.
\end{abstract}
\maketitle

\section{Introduction}
Machine Unlearning (MU), first introduced by \cite{Cao2015} refers to removing the influence of specific data points from machine learning (ML) models without retraining the model from scratch. This is important for several reasons. Privacy regulations, such as the European Union's (EU) General Data Protection Regulation (GDPR) require companies to delete user data on request from storage systems, as well as the influence of data points on model learning. Retraining models for each deletion is prohibitively expensive \cite{Nguyen2025SurveyMachineUnlearning}. 

Unlearning also becomes necessary for security reasons. Contemporary adversarial attacks can introduce carefully crafted, misleading inputs that distort model behaviour, such as forcing the model to output incorrect healthcare predictions. Additionally, where personalised systems adapt to user interests, an incorrect or accidental user action warrants immediate removal lest recommender systems continue to generate unwarranted suggestions  \cite{Nguyen2025SurveyMachineUnlearning}. ML models are also susceptible to training data bias, which enables them to discriminate against certain groups. \cite{wang2024machineunlearningcomprehensivesurvey} proposes the removal of certain attributes from the data that are more prone to induce bias.

Quantum Machine Learning (QML) refers to algorithms designed to run on Quantum Computers \cite{Sergioli2020}.  With the increased traction of Quantum Machine Learning (QML) in recent years, data privacy, security, and protection remain a persistent issue. \cite{Su2025Unlearning} design a Membership Inference Attack (MIA) tailored for QML models, demonstrating clear evidence of membership leakage from QML models. Their experiments validate Quantum Machine Unlearning (QMU) as a viable mechanism to resolve this issue. Considering the recent traction towards QML, there has been even less work in QMU. 

Unlearning is generally achieved by either forgetting a certain class (Class Unlearning), forgetting specific data points (Instance Unlearning), or forgetting features of the data (Feature Unlearning). \cite{Bourtoule, zagardo2024practicalapproachmachineunlearning, trippa2024nablataugradientbasedtaskagnostic}

Very recently, \cite{crivoi2025machineunlearningeraquantum} presented the first empirical study of MU in hybrid quantum--classical models, adapting a range of classical unlearning techniques to variational quantum circuits and proposing a complement-label strategy that enforces high-entropy predictions on forgotten samples. Their results demonstrate that quantum-enhanced models can support effective unlearning, with performance depending strongly on circuit depth and architectural choices.

In this work, we introduce a distribution-guided class unlearning framework for
variational quantum classifiers that remove the predictive influence of a
specified class without retraining from scratch.
Our approach formulates unlearning as a constrained optimization problem that
suppresses forgotten-class predictions while preserving the model’s behaviour on
retained data through soft anchor constraints.
Unlike existing unlearning methods that rely on uniform probability suppression
or full data deletion, we construct a data-driven forget target distribution that
redistributes probability mass according to the semantic similarity inferred from
the original model’s outputs.
The resulting objective admits a principled Lagrangian interpretation and is
optimized directly at the level of quantum circuit parameters using
parameter-shift gradients. We demonstrate that this approach achieves selective forgetting through
localized parameter updates, maintains retained-class structure, and is
effective across datasets of differing complexity, establishing a practical and theoretically grounded unlearning mechanism for near-term quantum machine
learning models.

\section{Theoretical Foundations}
Machine unlearning refers to the removal of data and its corresponding influence from a trained machine learning model, without requiring full retraining from scratch. Let $Z$ denote the space of all possible examples, and let $Z^*$ denote the set of all finite datasets drawn from $Z$. Given a dataset $D \in Z^*$, a learning algorithm $A : Z^* \rightarrow H$ maps $D$ to a hypothesis $A(D)$ in the hypothesis space $H$, which represents the set of admissible model parameters.

Given a trained model $A(D)$ and a subset of samples $D_f \subseteq D$ to be forgotten, an unlearning mechanism $U$ takes as input $(D, D_f, A(D))$ and produces an updated model intended to approximate the counterfactual model obtained by training without $D_f$:
\begin{equation}
    U(D_f, D, A(D)) \approx A(D \setminus D_f).
\label{eq:cmu}
\end{equation}
This formulation captures the central objective of MU: removing the functional influence of $D_f$ while preserving the model's performance on retained data.

\cite{Li2025Unlearning} proposes two formal definitions of unlearning, which we summarise below.

\subsection{Formal Definition 1}
Given a learning algorithm $A(\cdot)$, a dataset $D$, and a forgotten subset $D_f \subseteq D$, an unlearning algorithm $U(\cdot)$ is said to perform \emph{exact unlearning} if
\begin{equation}
\Pr\big(A(D \setminus D_f)\big) = \Pr\big(U(D, D_f, A(D))\big) = 1.
\label{eq:fomal_def1}
\end{equation}

\noindent This definition emphasizes behavioural equivalence rather than procedural equivalence. The unlearning algorithm is not required to retrain the model from scratch, but rather to produce an outcome that is indistinguishable from retraining on $D \setminus D_f$. The probability measure captures stochasticity arising from initialization, data ordering, and optimization randomness.

\subsection{Formal Definition 2}
More generally, given a hypothesis space $H$, an unlearning algorithm $U$ is said to perform exact unlearning if, for all measurable subsets $T \subseteq H$, datasets $D \in Z^*$, and forgotten subsets $D_f \subseteq D$,
\begin{equation}
    \Pr\big(A(D \setminus D_f) \in T\big)
    =
    \Pr\big(U(D, D_f, A(D)) \in T\big).
\label{eq:formal_def2}
\end{equation}

\noindent Here, $T$ represents a set of hypotheses satisfying certain structural or performance properties rather than a single parameter vector. Equality in distribution ensures that the unlearning procedure reproduces the same statistical behaviour as retraining, even if individual parameter realizations differ.

In practice, exact unlearning is often infeasible for modern models. Consequently, approximate unlearning techniques have been proposed, including variational Bayesian unlearning \cite{nguyen2020variationalbayesianunlearning}, which ensures that the posterior induced by unlearning remains close (in KL divergence) to the retrained posterior.

\section{Quantum Machine Unlearning}

QMU extends the classical unlearning paradigm to quantum and hybrid quantum-classical models. Unlike classical models parametrized in Euclidean space, quantum models are described by quantum states or density operators acting on Hilbert spaces. As a result, unlearning procedures must respect fundamental quantum constraints such as no-cloning and no-deletion.

\cite{Shaik2025QuantumUnlearning} formalizes QMU as a contraction of distinguishability between the model state before unlearning and the counterfactual state obtained by excluding the forgotten data. In this framework, an unlearning operation is modelled as a completely positive and trace-preserving (CPTP) map $\mathcal{E}$ acting on the model state $\rho$. CPTP maps represent the most general physically realizable quantum transformations.

The contractivity of quantum channels implies that distinguishability between quantum states cannot increase under $\mathcal{E}$. This is formalized by the quantum data-processing inequality:
\begin{equation}
\label{eq:cptp}
D(\rho \| \sigma) \ge D\big(\mathcal{E}(\rho)\,\|\,\mathcal{E}(\sigma)\big),
\end{equation}
where $D(\cdot\|\cdot)$ denotes a suitable quantum divergence measure. In the context of unlearning, this inequality motivates viewing forgetting as a contractive reduction of data-dependent distinguishability, consistent with physically admissible quantum evolutions. Since quantum information cannot be deleted outright, QMU redistributes data-dependent information to the environment while preserving physical consistency. Accordingly, for variational quantum classifiers, contractivity is reflected in reduced distinguishability at the level of observable predictive distributions.

An effective QMU method should therefore satisfy three criteria: \textbf{efficiency}, requiring limited computational overhead; \textbf{completeness}, ensuring removal of the forgotten data's influence; and \textbf{verifiability}, providing measurable evidence of forgetting.


\section{Methodology} 
\begin{figure}[t] \centering \includegraphics[width=0.8\linewidth]{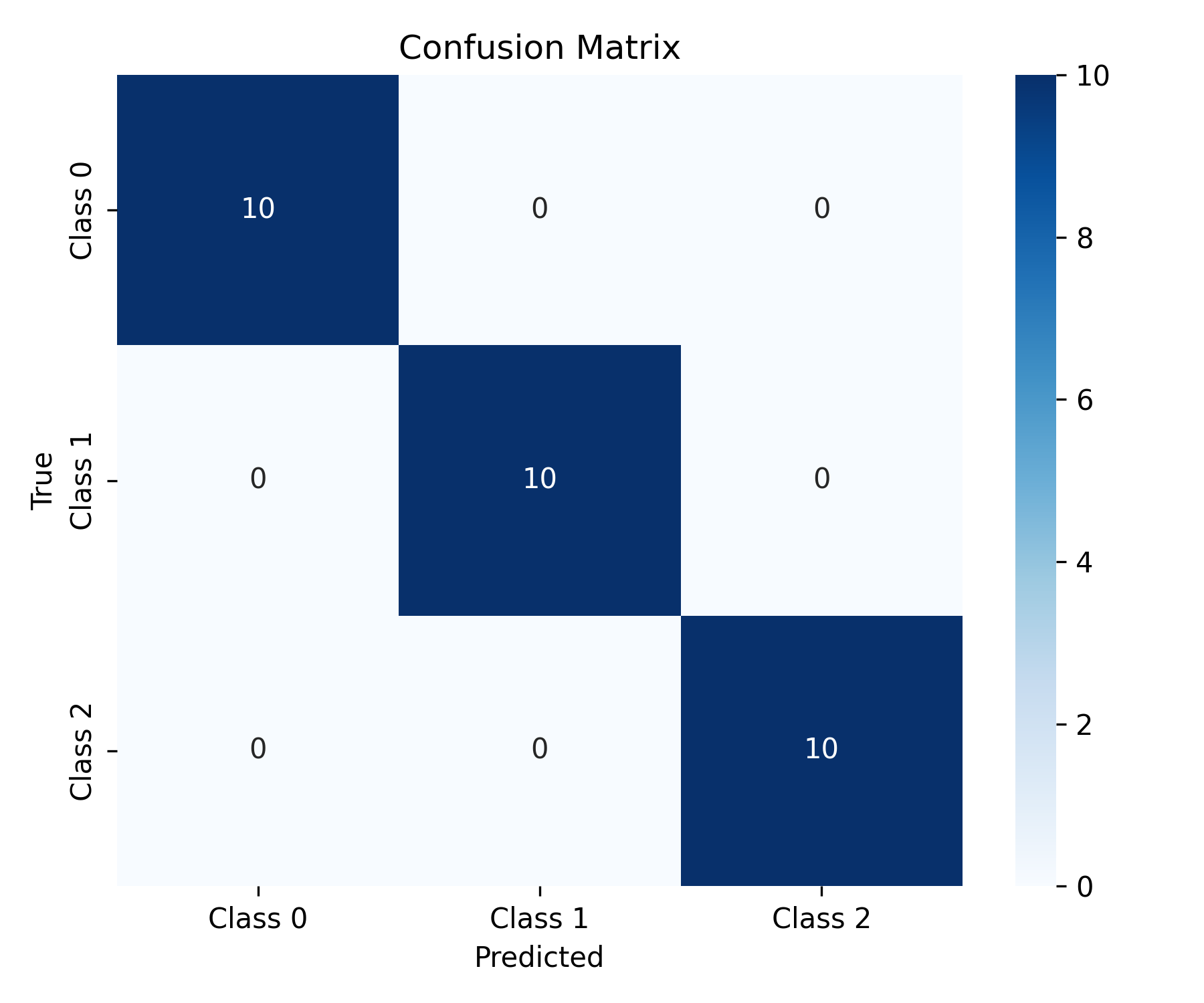} \caption{ Confusion matrix for the Iris dataset after training the variational quantum classifier. Rows correspond to true class labels and columns to predicted labels. The model achieves near-perfect separation across all three classes. } \label{fig:iris_learn_confusion} \end{figure}

\begin{figure}[t] \centering \includegraphics[width=0.8\linewidth]{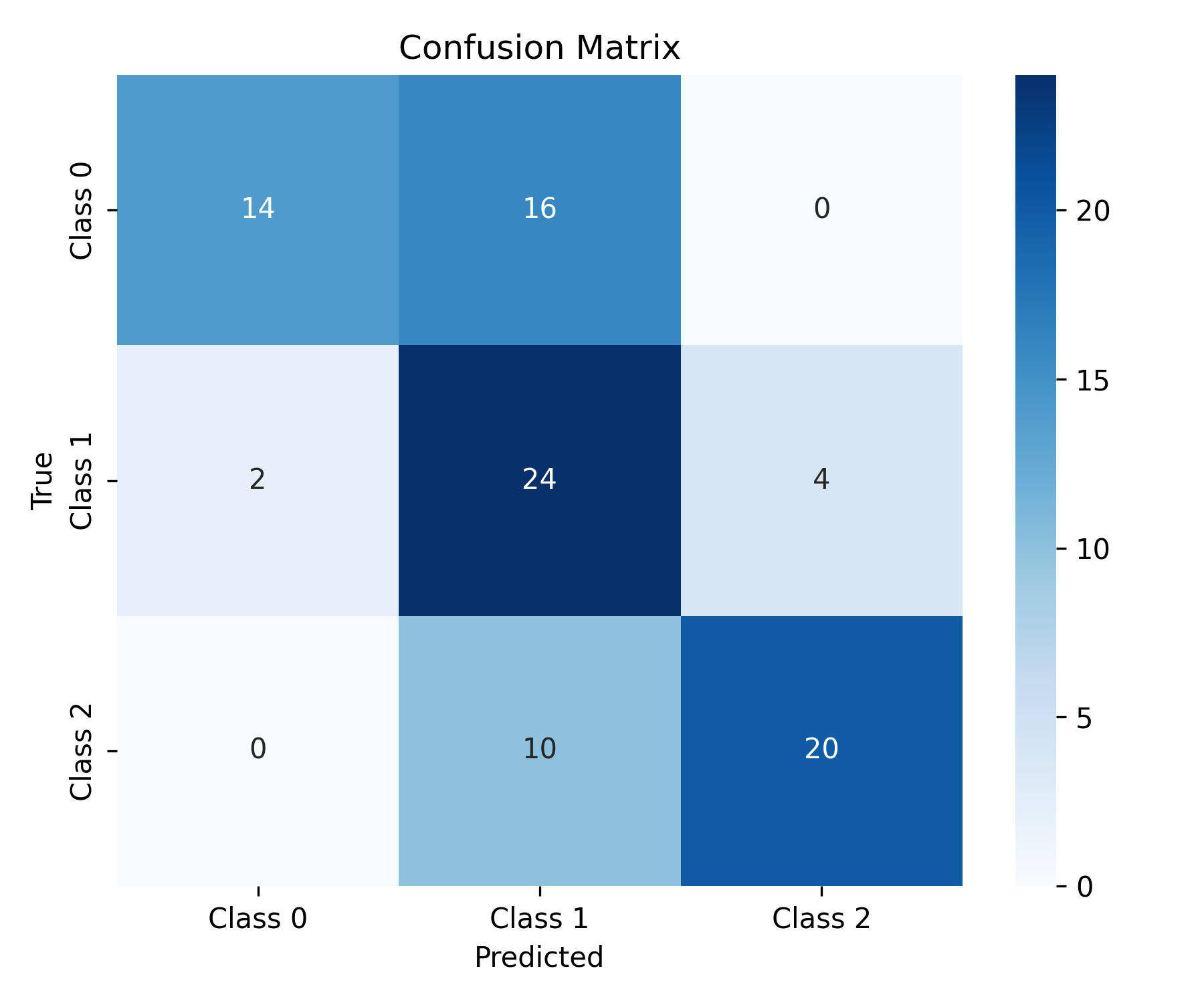} \caption{ Confusion matrix for the Covertype dataset (classes 3, 5, and 7) after training. While correct predictions dominate the diagonal, residual confusion between classes reflects the increased complexity of the dataset compared to Iris. } \label{fig:covertype_learn_confusion} \end{figure}

We consider supervised three-class classification with four continuous input features and present a distribution-guided unlearning framework applicable to this setting. We instantiate the framework on two representative classification problems — one low-complexity (Iris) and one higher-complexity (Covertype) — to illustrate how the same unlearning objective behaves across regimes. For Covertype, classes 3, 5, and 7 are selected (remapped to 0,1,2), and the original 54-dimensional inputs are reduced to four dimensions via principal component analysis. Each input, in this case $x \in \mathbb{R}^4$, is min--max scaled to $[0,\pi]$ and embedded into a six-qubit variational quantum circuit. Features are encoded on qubits 0--3 using single-qubit rotations $R_y(x_i)$, with pairwise feature interactions introduced via entangling blocks of the form $\mathrm{CX}(a,b)\,R_z(x_a x_b)\,\mathrm{CX}(a,b)$ applied to qubit pairs $(0,1)$, $(1,2)$, and $(2,3)$. The feature map is reuploaded twice in a layered feature map–ansatz–feature map–ansatz structure. The ansatz consists of ring entanglement with three repetitions. In each repetition, all six qubits undergo parametrised $R_y(\theta)$ and $R_z(\theta)$ rotations, followed by a ring of CNOT gates $\mathrm{CX}(q,(q+1)\bmod 6)$. This architecture results in 72 trainable parameters. Classification is performed by measuring Pauli-$Z$ expectation values on qubits 3, 4, and 5, which are treated as logits and passed through a softmax to produce class probabilities. Model parameters are trained by minimizing the mini-batch cross-entropy loss using stochastic gradient-based optimization. Gradients with respect to each circuit parameter are estimated via the parameter-shift rule \begin{equation} \frac{\partial L(w)}{\partial w_i} = \frac{L\!\left(w + \frac{\pi}{2} e_i\right) - L\!\left(w - \frac{\pi}{2} e_i\right)}{2}, \label{eq:param_shift} \end{equation} where $L(w)$ denotes the mini-batch loss and $e_i$ is the $i$-th standard basis vector. At each iteration, a mini-batch of 10 samples is drawn uniformly without replacement, and parameters are updated using Adam with a cosine learning-rate schedule over 300 iterations. Parameters are initialized with small Gaussian noise. The final model is selected by retaining the parameter configuration that achieves the lowest validation loss during training. Confusion matrices for Iris and Covertype classification after standard training are shown in Figures~\ref{fig:iris_learn_confusion} and~\ref{fig:covertype_learn_confusion}, respectively. \begin{figure}[t] \centering \includegraphics[width=\linewidth]{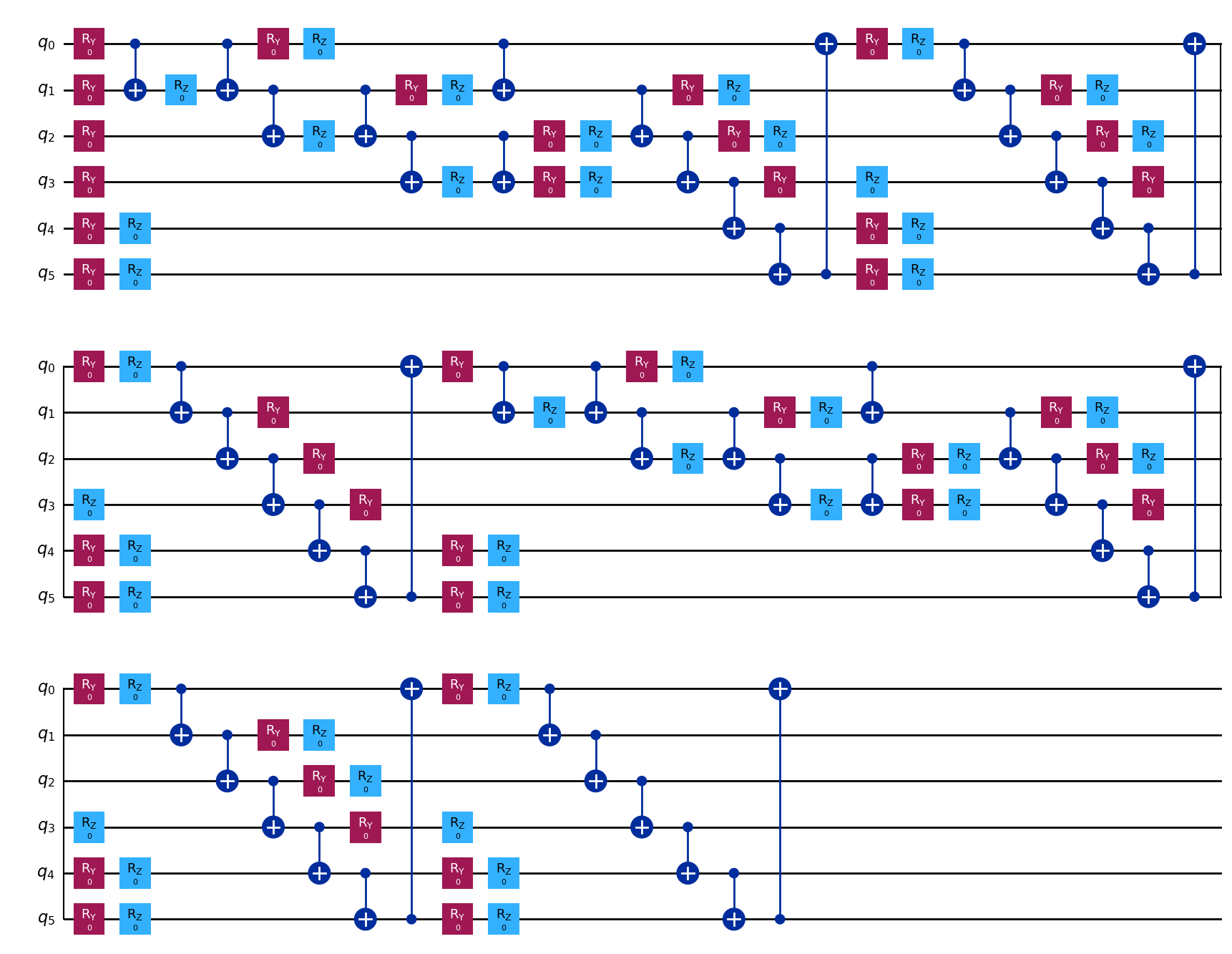} \caption{Variational quantum classifier architecture. A four-feature data-encoding map is reuploaded twice and interleaved with a hardware-efficient ansatz on six qubits. Features are embedded via single-qubit rotations and pairwise entangling blocks, followed by repeated layers of parametrized $R_y$–$R_z$ rotations and ring CNOT entanglement. Pauli-$Z$ expectation values on the readout qubits are used to produce class logits.} \label{fig:circuit} \end{figure}  

The formulation that follows depends only on the model’s predictive distributions and parameter space, and does not rely on dataset-specific structure beyond a fixed classification setting. The proposed unlearning objective is defined at the level of output distributions and model parameters, and is independent of the particular dataset, feature embedding, or circuit architecture used to instantiate the classifier.

\subsection{Distribution-Guided Class Unlearning}
\label{sec:unlearning}
Let a trained classifier with parameters $w_{\mathrm{orig}}$ output a
$K$-class predictive distribution $p_{w_{\mathrm{orig}}}(y\mid x)$.
Given a forgotten class $f$ and a forget set $F$ consisting of training samples with label $f$,
our goal is to compute updated parameters $w$ such that:
(i) for inputs $x\in F$, the unlearned model assigns negligible probability to the forgotten class,
i.e., $p_{w}(f\mid x)$ is suppressed; and
(ii) on the retained data, the model's predictions remain close to the original model
(and ideally close to a {gold retrained} model trained from scratch without $F$).
In the following, we operationalize these objectives using a divergence-based forget loss on $F$
and a constraint that preserves behaviour on a retained anchor set.

After standard training, we perform targeted class unlearning to remove predictive influence
associated with a specified class $f$.
To this end, the data are partitioned into a \emph{forget set} $F$,
containing samples associated with class $f$,
and an \emph{anchor set} $A$ consisting of retained samples whose predictive behaviour
should be preserved.

\subsubsection{Distribution-Guided Forget Target}
\label{sec:forget_target}
To guide the redistribution of probability mass during unlearning,
we define a fixed target distribution $q \in \Delta^{K-1}$ satisfying $q_f = 0$.
The remaining probability mass is distributed across non-forgotten classes
according to the original model's predictions on a calibration subset
$S \subseteq F$ of the forget class.
Specifically, we compute the mean softmax output of the original model on $S$
and assign
\begin{equation}
\label{eq:forget-target}
q_k \propto
\left(
\mathbb{E}_{x \in S}\!\left[p_{w_{\mathrm{orig}}}(k \mid x)\right]
\right)^{\beta},
\qquad k \neq f,
\end{equation}
followed by normalization.
The exponent $\beta > 0$ controls the sharpness of the target distribution. 
This similarity-based construction redistributes probability mass according to
the original model’s output distribution on the forget class, thereby favouring
target classes that the model already considers similar.
Consequently, probability mass is reassigned toward semantically proximate
outputs rather than being uniformly distributed across all non-forgotten classes.
Empirical evidence demonstrating the limitations of uniform redistribution,
particularly in higher-complexity settings, is provided in
~\ref{app:lagrangian_proof}.
Because the target distribution in \ref{eq:forget-target} is constructed solely from the model’s output probabilities on the forgotten class, the same procedure applies directly to any three-class classification task with continuous features, independent of the underlying dataset.
\subsubsection{Anchor Reference Distributions}
\label{sec:anchor_ref}
To preserve the model's behaviour on retained data,
we cache reference distributions
$p_{w_{\mathrm{orig}}}(\cdot \mid x)$
for all $x \in A$.
Using soft reference distributions enables preservation of the relative class
similarities and predictive calibration on anchor samples.

\subsubsection{Unlearning Objective}
\label{sec:unlearning_objective}
Unlearning is performed by maximizing the following objective:
\begin{equation}
\label{eq:unlearning_objective}
\begin{aligned}
J(w)
&=
\mathbb{E}_{x \in F}
\left[
\sum_{k=1}^{K} q_k \log p_w(k \mid x)
\right]
\\
&\quad+
\alpha
\mathbb{E}_{x \in A}
\left[
\sum_{k=1}^{K} p_{\mathrm{w_{orig}}}(k \mid x)\log p_w(k \mid x)
\right]
\\
&\quad-
\lambda \| w - w_{\mathrm{orig}} \|_2^2 ,
\end{aligned}
\end{equation}
where $\alpha \ge 0$ and $\lambda \ge 0$ are hyper-parameters. The first term encourages the model's predictions on the forget set
to align with the similarity-guided target distribution $q$,
which assigns zero probability to the forgotten class.
The second term preserves the original model's predictive behaviour
on anchor samples, while the final term penalizes large deviations
from the original parameters.

\subsubsection{Optimization and Interpretation}
\label{sec:unlearning_opt}
The objective in~\eqref{eq:unlearning_objective} is maximized via gradient ascent,
using parameter-shift gradient estimates and optional mini-batching over
the forget and anchor sets.
A theoretical interpretation of this objective as a Lagrangian relaxation
of a constrained unlearning problem is provided in Theorem \ref{thm:langragian_unlearning},
with proof deferred to \ref{app:lagrangian_proof}.

\begin{theorem}[Lagrangian formulation of distribution-guided unlearning]
\label{thm:langragian_unlearning}
Let $p_w(\cdot\mid x)$ denote the model softmax distribution induced by parameters $w$.
Let $F$ be a forget set and $A$ be an anchor (retained) set.
Let $p_{\mathrm{ref}}(\cdot\mid x) := p_{w_{\mathrm{orig}}}(\cdot\mid x)$ be the reference distribution
produced by the original parameters $w_{\mathrm{orig}}$ on anchor samples.
Let $q \in \Delta^{K-1}$ be a fixed target distribution satisfying $q_f = 0$ for the forgotten class $f$.

Define
\begin{align*}
\mathcal{L}_F(w)
&= \frac{1}{|F|} \sum_{x \in F} \sum_{k=1}^K q_k \log p_w(k \mid x), \\
\mathcal{L}_A(w)
&= \frac{1}{|A|} \sum_{x \in A} \sum_{k=1}^K
p_{\mathrm{ref}}(k \mid x)\log p_w(k \mid x).
\end{align*}

\noindent Then, maximizing the objective
\begin{equation}
\label{eq:final_objective}
J(w)
=
\mathcal{L}_F(w)
+
\alpha\,\mathcal{L}_A(w)
-
\lambda \| w - w_{\mathrm{orig}} \|_2^2,
\quad \alpha,\lambda \ge 0,
\end{equation}
is equivalent, up to additive constants independent of $w$, to maximizing the Lagrangian relaxation of the constrained optimization problem
\begin{equation}
\label{eq:final_constrained_problem}
\begin{aligned}
\max_{w} \quad
& \mathcal{L}_F(w) \\
\rm{such~that} \quad
& \frac{1}{|A|} \sum_{x \in A}
\mathrm{KL}\!\left(
p_{\mathrm{ref}}(\cdot \mid x)
\,\|\, p_w(\cdot \mid x)
\right)
\le \varepsilon, \\
& \| w - w_{\mathrm{orig}} \|_2^2 \le \rho,
\end{aligned}
\end{equation}
for some $\varepsilon,\rho \ge 0$.
\end{theorem}
\subsection{Implementation Details}
All experiments were implemented using Qiskit and executed on a noiseless statevector quantum simulator. Variational quantum classifiers were constructed using six qubits with data reuploading and a hardware-efficient ansatz composed of parametrised single-qubit rotations and nearest-neighbour entangling gates. Model outputs were obtained via exact expectation-value evaluation of Pauli-Z measurements on designated readout qubits, without shot noise or device-level errors. Gradients were computed using the parameter-shift rule, and optimization was performed using the Adam optimizer. No hardware noise models were assumed, as the focus of this work is on the algorithmic behaviour of quantum machine unlearning rather than device-specific performance.

\subsection{Results}
\begin{figure}[t]
\centering
\includegraphics[width=0.8\linewidth]{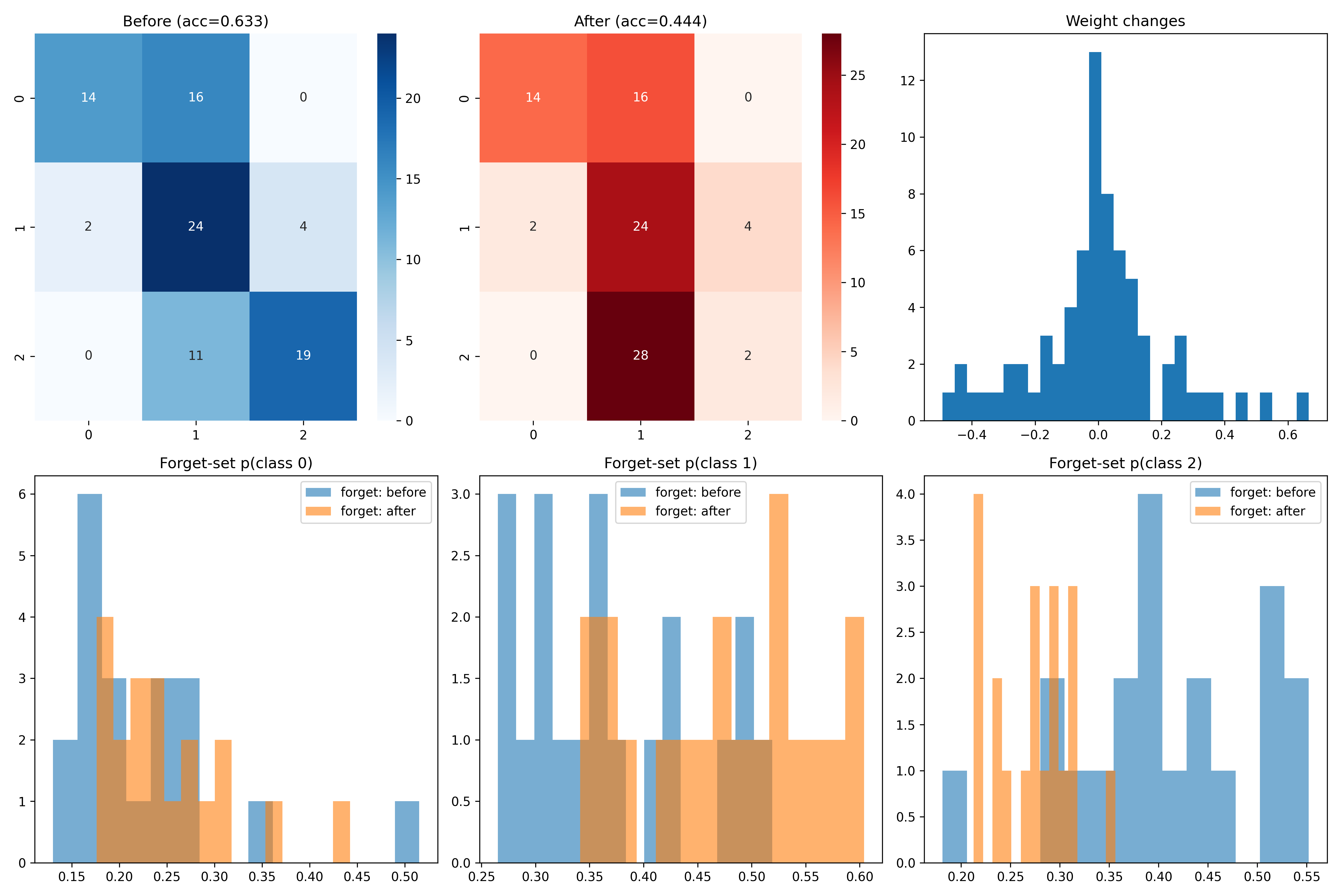}
\caption{
Confusion matrices before and after targeted unlearning on the Covertype dataset
with class~2 forgotten. Predictions assigned to the forgotten class are strongly
suppressed after unlearning, with errors primarily redistributing to a single
retained class.
}
\label{fig:covertype_unlearning}
\end{figure}

\begin{figure}[t]
\centering
\includegraphics[width=0.8\linewidth]{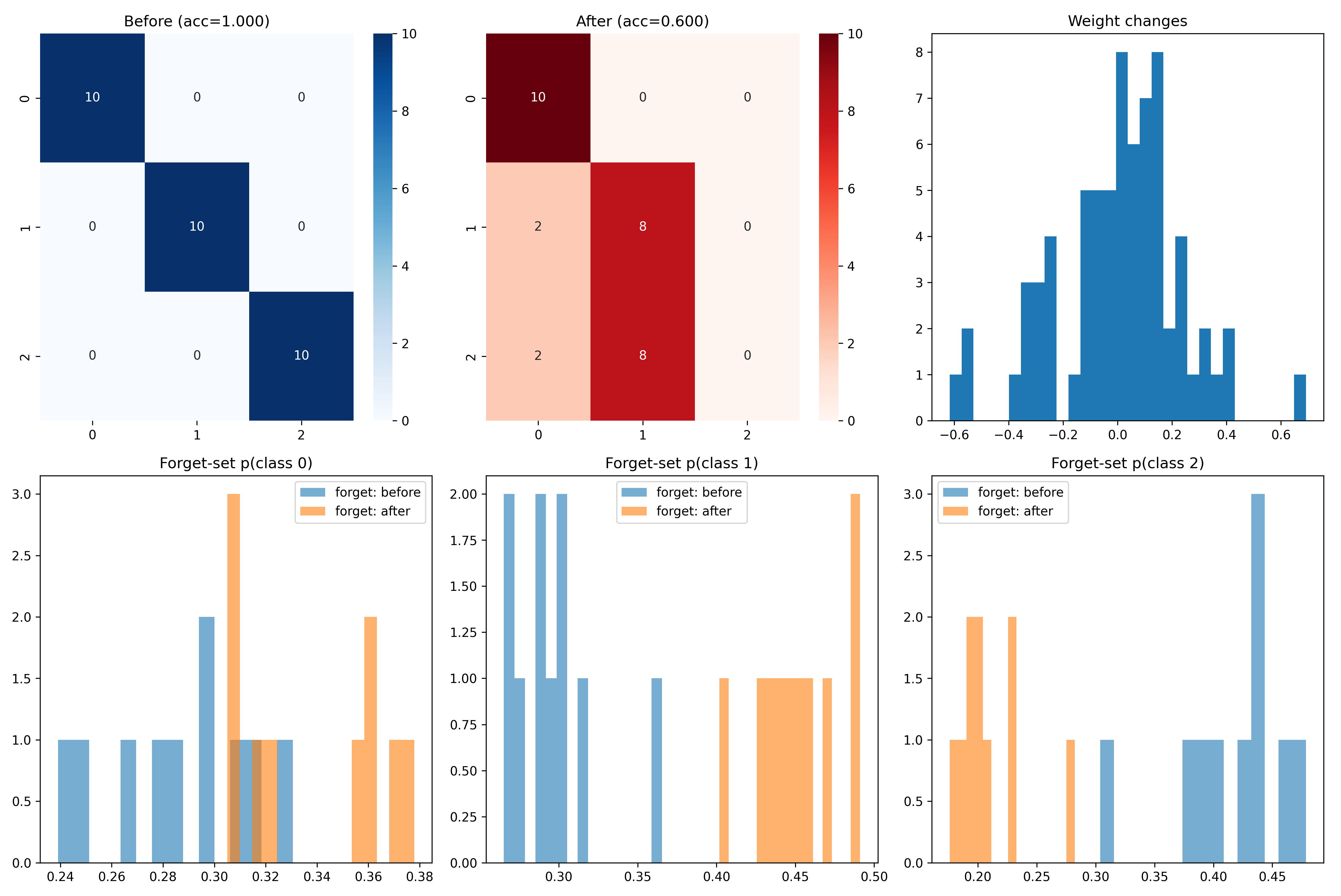}
\caption{
Confusion matrices before and after targeted unlearning on the Iris dataset
with class~2 forgotten. Forgotten-class predictions are eliminated, while retained
classes largely preserve pre-unlearning behaviour.
}
\label{fig:iris_unlearning}
\end{figure}

Unless stated otherwise, all unlearning experiments use 
$\alpha=1.0$  and $\lambda = 0.01$ reflecting equal weighting of the forgetting and preservation terms, together with a mild quadratic anchoring to the original parameters.
Targeted unlearning produces a sharp and selective degradation in performance for
the forgotten class, while largely preserving retained-class behaviour, as shown  in Figures~\ref{fig:covertype_unlearning} and \ref{fig:iris_unlearning}.
On Covertype, recall for the forgotten class decreases from $0.633$ to $0.067$,
while recall for the two retained classes remains unchanged at $0.467$ and $0.800$,
respectively.
The mean predicted probability assigned to the forgotten class on its own data
drops from $0.4055$ to $0.2715$, indicating a substantial reduction in residual
confidence.
For Iris, forgotten-class recall decreases from $1.000$ to $0.000$, while recall
for class~0 remains at $1.000$ and class~1 decreases modestly from $1.000$ to $0.800$.
The corresponding forgotten-class mean probability decreases from $0.4155$ to $0.2098$.
Across both datasets, post-unlearning confusion matrices show that errors introduced
by unlearning primarily redistribute forgotten samples to a single retained class,
consistent with similarity-guided probability reassignment rather than
indiscriminate suppression.


\onecolumn

\twocolumn

Beyond accuracy-based metrics, we evaluate how closely the unlearned model
approximates true retraining by comparing its predictive distributions to those of a
gold retrained model trained on retained data only. We compute the Kullback--Leibler (KL) divergence between the gold retrained and
unlearned predictive distributions on retained-class test samples. Probabilities
are renormalized over retained labels to exclude trivial effects from probability
mass assigned to the forgotten class. This ensures that divergence reflects
agreement on retained-class behaviour rather than the degree of forgetting itself.
This comparison serves as an empirical proxy for distributional equivalence in
Definition~2, assessing whether unlearning reproduces the statistical behaviour of
retraining on retained data.

On the Covertype dataset, the unlearned model achieves a mean KL divergence of
$0.047 \pm 0.047$ relative to the gold retrained model, with a median of $0.034$.
This low divergence indicates that, despite aggressive suppression of the
forgotten class, the unlearned model closely matches the behaviour of a model
trained from scratch on retained data.
At the same time, the unlearned model assigns a larger fraction of probability
mass to the forgotten class than the gold-retrained model on retained samples
($0.295$ vs.\ $0.195$), reflecting a mild under-learning that does not
substantially affect retained-class predictions. While our experiments focus on
class unlearning, extending similarity-guided and constraint-based formulations to
instance-level unlearning remains an important direction for future work.


\begin{table}[t]
\centering
\caption{Class-wise recall and forget-class confidence before and after unlearning.
The forgotten class is class~2 in all experiments.}
\label{tab:classwise_results}
\footnotesize
\begin{tabular}{llccc}
\toprule
Dataset & Class & Recall (B) & Recall (A) & $\Delta$ Recall \\
\midrule
\multirow{3}{*}{Covertype}
& 0 & 0.467 & 0.467 & 0.000 \\
& 1 & 0.800 & 0.800 & 0.000 \\
& 2 (F) & 0.633 & 0.067 & $-0.566$ \\
\midrule
\multirow{3}{*}{Iris}
& 0 & 1.000 & 1.000 & 0.000 \\
& 1 & 1.000 & 0.800 & $-0.200$ \\
& 2 (F) & 1.000 & 0.000 & $-1.000$ \\
\bottomrule
\end{tabular}

\vspace{0.3em}
\raggedright
\footnotesize
\textit{Forget-class mean probability:}
Covertype $0.4055 \rightarrow 0.2715$,
Iris $0.4155 \rightarrow 0.2098$.
\end{table}

\begin{table}[t]
\centering
\caption{Divergence between gold retrained and unlearned models on retained test data. KL divergence computed after renormalizing over retained labels only. Lower values indicate closer agreement.}
\label{tab:kl_to_gold}
\footnotesize
\begin{tabular}{@{}lcccc@{}}
\toprule
Dataset & Ret. Labels & Mean KL & Med. KL & Max KL \\
\midrule
Covertype & \{0,1\} & 0.0468 & 0.0342 & 0.1625 \\
\bottomrule
\end{tabular}

\vspace{0.3em}
\raggedright
\footnotesize
\textit{Mean prob.\ on forgotten class (retained samples):}
Gold 0.195, Unlearned 0.295.
\end{table}

\section{Conclusions}
We introduced a generalised four-feature three-class distribution-guided and constrained framework for quantum machine unlearning that selectively suppresses forgotten-class influence while preserving retained behaviour. Experiments on variational quantum classifiers demonstrate sharp forgetting, minimal degradation on retained classes, and close alignment with gold retraining baselines. From a quantum information perspective, our framework is motivated by physical admissibility constraints on unlearning operations. In practice, we realize these principles through constrained variational optimization, with unlearning quality evaluated empirically via alignment to gold retraining.

\section*{Data Availability}
All data generated or analysed during this study are included in this published article and the accompanying figures. The code supporting the findings of this study will be made publicly available at
\href{https://github.com/nausherwan-malik/Distribution-Guided-Constrained-Quantum-Machine-Unlearning}
{\nolinkurl{github.com/nausherwan-malik/Distribution-Guided-Constrained-Quantum-Machine-Unlearning}}.

\section*{Author Contributions}
NM designed and implemented the proposed methodology; conducted all experiments and data analysis; and wrote the draft of the manuscript. MF proposed the research idea, supervised the project and data analysis, and guided the writing of the manuscript. ZK supervised the research and provided oversight and feedback on the study. All authors reviewed and approved the final version of the manuscript.



\bibliography{references}

\clearpage
\onecolumn
\appendix
\section{Limitations of uniform forget-target assignment}
\label{uniform_covertype}
\begin{figure}[h]
    \centering
    \includegraphics[width=0.8\linewidth]{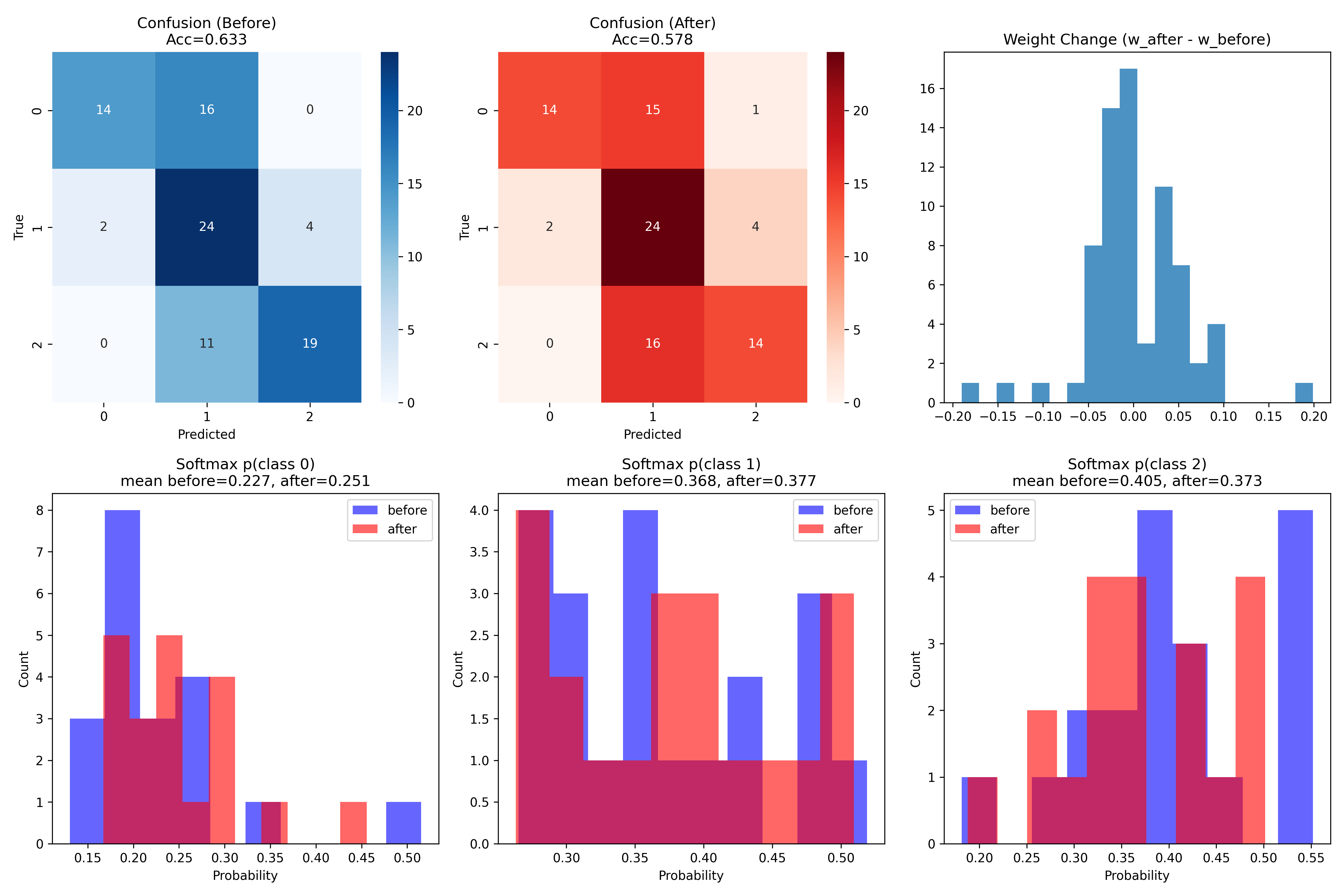}
    \caption{
Effect of uniform forget-target assignment on the Covertype dataset.
The confusion matrices show model predictions before (left) and after (center)
unlearning using a uniform target distribution over non-forgotten classes, while
the right panel reports the distribution of parameter changes.
Uniform redistribution reduces, but does not eliminate, predictions assigned to
the forgotten class, with the mean forgotten-class probability decreasing from
$0.405$ to $0.373$.
Probability mass is dispersed across retained classes rather than concentrated on
a semantically similar class, resulting in less structured post-unlearning
confusion patterns compared to similarity-guided unlearning.}
    \label{fig:uniform_unlearning_covertype}
\end{figure}

Figure~\ref{fig:uniform_unlearning_covertype} illustrates the effect of assigning a
uniform target distribution over non-forgotten classes during unlearning on the
Covertype dataset.
While uniform redistribution reduces the model’s confidence in the forgotten
class, the suppression is substantially weaker than under similarity-guided
targets, with the mean forgotten-class probability decreasing only from $0.405$
to $0.373$.
Class-wise confusion matrices show that a large fraction of forgotten samples
continue to be predicted as the forgotten class after unlearning, and the overall
accuracy remains relatively high at $0.578$, indicating incomplete forgetting.
Moreover, probability mass is redistributed diffusely across retained classes
rather than concentrating on a semantically proximate class, leading to less
structured confusion patterns.
These results highlight that uniform target assignment fails to sufficiently
remove class-specific predictive influence in more complex datasets, motivating
the use of similarity-informed targets.

\section{Proof of Theorem 1}
\label{app:lagrangian_proof}

For any anchor sample $x \in A$,
\[
\sum_k p_{\mathrm{ref}}(k \mid x)\log p_w(k \mid x)
=
-\mathrm{KL}\!\left(
p_{\mathrm{ref}}(\cdot \mid x)
\,\|\, p_w(\cdot \mid x)
\right)
+
\sum_k p_{\mathrm{ref}}(k \mid x)\log p_{\mathrm{ref}}(k \mid x),
\]
where the second term does not depend on $w$.
Averaging over $x \in A$ yields
\[
\mathcal{L}_A(w)
=
-
\frac{1}{|A|} \sum_{x \in A}
\mathrm{KL}\!\left(
p_{\mathrm{ref}}(\cdot \mid x)
\,\|\, p_w(\cdot \mid x)
\right)
+ C,
\]
for a constant $C$ independent of $w$. Substituting this expression into~\eqref{eq:final_objective} and discarding constants independent of $w$ gives
\[
J(w)
\equiv
\mathcal{L}_F(w)
-
\alpha
\frac{1}{|A|} \sum_{x \in A}
\mathrm{KL}\!\left(
p_{\mathrm{ref}}(\cdot \mid x)
\,\|\, p_w(\cdot \mid x)
\right)
-
\lambda \| w - w_{\mathrm{orig}} \|_2^2.
\]

\noindent This expression is precisely the Lagrangian associated with the constrained optimization problem
in~\eqref{eq:final_constrained_problem}, with Lagrange multipliers $\alpha$ and $\lambda$, completing the proof.

\section{Ablation Studies}
\label{app:ablations}

This appendix investigates the sensitivity of the proposed distribution-guided
quantum machine unlearning framework to its principal design choices and
hyperparameters. Given that the Iris dataset is a simple, low-complexity classification task,  
all ablations are conducted on the Covertype dataset using the variational quantum
classifier described in Section~\ref{sec:unlearning}, unless stated otherwise. 
Unless varied explicitly, the default unlearning parameters are
$\alpha = 1.0$, $\lambda = 0.01$, and $\beta = 1.0$.
To avoid repeating the class unlearning discussed in the previous discussions, we exclude unlearning in class 2 in these experiments. 

Throughout this appendix, forgetting effectiveness is quantified by the mean
predicted probability assigned to the forgotten class on its own samples,
while retained utility is measured by test accuracy on all classes.

\subsection{Class-wise forgetting robustness}
\label{app:classwise}

We first evaluate whether the proposed method depends on the specific choice of
forgotten class.
Using the default unlearning configuration, we apply class-level unlearning to
each non-trivial class in turn and report the resulting forgetting strength and
retained accuracy.

\begin{table}[h]
\centering
\caption{Class-wise unlearning performance on Covertype.
Results are reported for non-trivial forgotten classes.
Lower forgotten-class probability indicates stronger forgetting.}
\label{tab:classwise_ablation}
\footnotesize
\begin{tabular}{cccc}
\toprule
Forgotten Class & Test Acc. (Before) & Test Acc. (After) & $p_f$ (After) \\
\midrule
0 & 0.633 & 0.411 & 0.260 \\
1 & 0.633 & 0.411 & 0.261 \\
\bottomrule
\end{tabular}
\end{table}

Across forgotten classes, the method consistently suppresses confidence assigned
to the forgotten class while maintaining comparable retained accuracy.
This indicates that the unlearning behaviour is not specific to a single class
choice and is robust across non-trivial forgetting scenarios.

\subsection{Sensitivity to similarity sharpness $\beta$}
\label{app:beta_ablation}

The distribution-guided target distribution in
Eq.~\eqref{eq:forget-target} depends on the sharpness parameter $\beta$,
which controls how strongly the probability mass is concentrated on the most
semantically similar retained class.
We sweep $\beta \in \{0.25, 0.5, 0.75, 1.0\}$ while keeping all other parameters fixed.

\begin{table}[h]
\centering
\caption{Effect of varying the similarity sharpness parameter $\beta$
on forgetting and retention.}
\label{tab:beta_ablation}
\footnotesize
\begin{tabular}{cccc}
\toprule
$\beta$ & Test Acc. (After) & $p_f$ (After) \\
\midrule
0.25 & 0.400 & 0.269 \\
0.50 & 0.411 & 0.267 \\
0.75 & 0.422 & 0.264 \\
1.00 & 0.411 & 0.261 \\
\bottomrule
\end{tabular}
\end{table}

Performance varies smoothly with $\beta$, with no sharp degradation in either
forgetting or retaining accuracy.
This indicates that the similarity-guided target construction is robust to
moderate changes in sharpness, and does not require precise tuning.

\subsection{Anchor set size sensitivity}
\label{app:anchor_size}

We next examine how the size of the retained anchor set affects unlearning
performance.
The anchor set is subsampled in a stratified manner to preserve class balance,
using fractions $\{10\%, 25\%, 50\%, 100\%\}$ of the original anchor data.

\begin{table}[h]
\centering
\caption{Effect of anchor set size on unlearning performance.}
\label{tab:anchor_size_ablation}
\footnotesize
\begin{tabular}{cccc}
\toprule
Anchor Fraction & Test Acc. (After) & $p_f$ (After) \\
\midrule
0.10 & 0.356 & 0.250 \\
0.25 & 0.411 & 0.260 \\
0.50 & 0.422 & 0.260 \\
1.00 & 0.411 & 0.261 \\
\bottomrule
\end{tabular}
\end{table}

\noindent Retention performance improves rapidly as the anchor fraction increases from
$10\%$ to $25\%$ and stabilizes thereafter, while forgetting effectiveness remains
largely unchanged.
This suggests that only a modest retained anchor subset is sufficient to preserve
model behaviour, highlighting the practical efficiency of the proposed constraint.

\subsection{Effect of anchor-preservation weight $\alpha$}
\label{app:alpha_ablation}

The coefficient $\alpha$ in Eq.~\eqref{eq:unlearning_objective} controls the
strength of the anchor-preservation term.
To assess its role, we vary $\alpha \in \{0, 1, 2\}$ while keeping all other
parameters fixed.

\begin{table}[h]
\centering
\caption{Effect of varying $\alpha$ on forgetting and retention.}
\label{tab:alpha_ablation}
\footnotesize
\begin{tabular}{cccc}
\toprule
$\alpha$ & Test Acc. (After) & $p_f$ (After) \\
\midrule
0 & 0.300 & 0.214 \\
1 & 0.411 & 0.261 \\
2 & 0.400 & 0.270 \\
\bottomrule
\end{tabular}
\end{table}

Removing the anchor constraint ($\alpha = 0$) leads to aggressive forgetting but
substantial degradation of retained accuracy.
Increasing $\alpha$ moderates forgetting while improving retention, consistent
with the Lagrangian interpretation of
Theorem~\ref{thm:langragian_unlearning}.

\subsection{Effect of parameter anchoring $\lambda$}
\label{app:lambda_ablation}

Finally, we study the effect of the quadratic parameter anchoring term in
Eq.~\eqref{eq:unlearning_objective} by varying
$\lambda \in \{0, 0.01, 0.1\}$.

\begin{table}[h]
\centering
\caption{Effect of varying the parameter anchoring coefficient $\lambda$.}
\label{tab:lambda_ablation}
\footnotesize
\begin{tabular}{cccc}
\toprule
$\lambda$ & Test Acc. (After) & $p_f$ (After) \\
\midrule
0 & 0.411 & 0.251 \\
0.01 & 0.411 & 0.261 \\
0.10 & 0.433 & 0.334 \\
\bottomrule
\end{tabular}
\end{table}

\noindent Larger values of $\lambda$ improve retention at the cost of weaker forgetting,
while $\lambda = 0$ retains effective forgetting but provides less stability.
This confirms that parameter anchoring acts as a secondary stabilization mechanism,
complementing the anchor distribution constraint rather than replacing it.

\paragraph{Summary}
Across all ablations, the proposed distribution-guided and constrained unlearning
framework exhibits stable and interpretable behaviour.
The method is robust to the choice of forgotten class, insensitive to moderate
changes in the similarity sharpness parameter, and requires only a modest anchor
subset to preserve retained behaviour.
The effects of $\alpha$ and $\lambda$ align closely with the Lagrangian
interpretation of the objective, providing empirical validation of the theoretical
formulation in Section~\ref{sec:unlearning}.

\end{document}